\documentclass[11pt,a4paper]{article}
\usepackage[hyperref]{acl2018}
\usepackage{times}
\usepackage{latexsym}

\usepackage{url}
\usepackage{graphicx,amsmath,amssymb,threeparttable,url,subfig,placeins} 
\usepackage{algorithm} 
\usepackage{algpseudocode} 

\usepackage{listings}
\usepackage{xcolor}
\usepackage{varwidth}
\usepackage{cprotect}

\aclfinalcopy 

\onecolumn

\title{CytonRL: an Efficient Reinforcement Learning\\
Open-source Toolkit Implemented in C++
}

\author{Xiaolin Wang\\
National Institute of Information and Communications Technology, Japan\\
{\tt xiaolin.wang@nict.go.jp }
}

\date{}

\begin{document}

\maketitle
\begin{abstract}
This paper presents an open-source enforcement learning toolkit named CytonRL~\footnote{https://github.com/arthurxlw/cytonRL}. The toolkit implements four recent advanced deep Q-learning algorithms from scratch using C++ and NVIDIA's GPU-accelerated libraries.  The code is simple and elegant, owing to an open-source general-purpose neural network library named CytonLib. Benchmark shows that the toolkit achieves competitive performances on the popular Atari game of Breakout.  
\end{abstract}

\section{Introduction}

Reinforcement learning (RL) is self learning what to do under an environment, in other words, how to map situations to actions, so as to maximize a numerical reward signal~\cite{sutton1998reinforcement}. RL is an meaningful artificial intelligence task, and will be extremely useful if it works. However, traditional real-world RL systems were usually built upon hand-crafted features from raw sensor data, which is a bottleneck of their performance. Therefore, 
learning to control agents directly from high-dimensional sensory inputs was considered  as one of the long-standing challenges of RL.

 Recently, the deep learning community has developed  deep neural networks to automatically extract high-level features from raw sensory data, leading to breakthroughs in computer vision~\cite{lecun1998gradient,krizhevsky2012imagenet,farabet2013learning, sermanet2013pedestrian,mnih2013machine} and speech recognition~\cite{povey2014parallel,dahl2012context,graves2013speech}. Excitingly, the RL community integreted this technology into their systems, and achieved the long-standing challenge~\cite{mnih2013play,mnih2015human}.

The breakthrough in RL will undoubtedly give birth to impressive progress in the related fields such as natural language processing and robotics. Therefore, we develop the open-source  toolkit of CytonRL in the hope to benefit research communities as well as industries.  

CytonRL is an open-source toolkit of deep Q-learning. It achieves competitive performances on the test environment of Atari 2600 test-bed~\cite{bellemare2013arcade} through following the works as,
\begin{description}
\item{Deep Q-Network}(DQN) proposed by \cite{mnih2013play} and \cite{mnih2015human};
\item{Double DQN} proposed by \cite{hasselt2015deep};
\item{Prioritized Experience Replay} proposed by \cite{schaul2015prioritized};
\item{Dueling DQN} proposed by \cite{wang2015dueling}.
\end{description}
In addition, the parameter settings of CytonRL has been carefully tuned for both efficiency and effectiveness.

CytonRL is built from scratch using C++ and NVIDIA's GPU-accelerated libraries, sharing the same strategy as the neural machine translation toolkit of CytonMT~\cite{wang2018cytonmt}. The advantages of CytonRL includes,
\begin{description}
\item[Running Efficiency] through better exploiting the power of GPU compared to the toolkit implemented in other languages, since C++ language is the genuine official language of NVIDIA -- the manufacturer of the GPU hardware;
\item[Code Simplicity] owing to an C++ open-source general-purpose neural network library   named CytonLib which is shipped as part of the source code.
\item[Programming Flexibility] as all low-level operations are visible to users. 
\end{description}

The following of this paper is organized as: the section~\ref{sec:method} descirbes the methods used in the toolkit of CytonRL; the section~\ref{sec:implementation} explains the implmentation; the section~\ref{sec:benchmark} presents the benchmark; the section~\ref{sec:conclusion} concludes this paper.


\section{Method}
\label{sec:method}
CytonRL has implemented four recent advanced deep Q-learning algorithms proposed by \cite{mnih2013play,hasselt2015deep,schaul2015prioritized,wang2015dueling}. The following subsections first introduce the background knowledge of reinforcement learning, and then present the details of these four algorithms.

\subsection{Background}
Suppose an agent interacts with an environment $\mathcal{E}$ in a sequence of actions, observations, and rewards~\citep{mnih2013play}. At each time-step,the agent selects an action $a_t$ from the set of legal game actions, $ A =\{1, \ldots, K\}$. The action is passed to  $\mathcal{E}$ and modifies its internal state. The agent both receives an reward $r_t$ and makes an new observation $x_{t+1}$ from $\mathcal{E}$.

Most often the observation $x_t$ does not fully specify the internal state of $\mathcal{E}$. Therefore, the sequence of actions and observations $s_t = x_1, a_1, x_2, a_2,\ldots,a_{t-1},x_t$ are considered as an representation of $\mathcal{E}$'s state, upon which strategies are learned.

The goal of the agent is to interact with $\mathcal{E}$ by selecting actions in a way that maximizes future rewards. There is an standard assumption that future rewards are discounted by a factor of $\gamma$ per time-step, as $\mathcal{E}$ is generally stochastic. The future discounted return at the time $t$ is defined as
\begin{equation}
R_t = \sum_{t'=t}^T \gamma^{t'-t}r_{t'},
\end{equation}
where $T$ is the time-step at which $\mathcal{E}$ decides to terminate.

In order to deduce the optimal policy, a helper function named optimal action-value function is defined as the maximum expected return achievable by following any strategy, after seeing some sequence $s$ and then taking some action $a$, formulated as,
\begin{equation}
Q^*(s,a) = \mathop{max}_\pi \mathbb{E} [R_t | s_t=s, a_t = a, \pi], 
\end{equation}
where $\pi$ is a policy mapping a sequence to actions or distributions over actions.

The optimal policy can be derived after knowing $Q^*(s,a)$, formulated as,
\begin{equation}
a^*(s) = \mathop{argmax}_a Q^*(s,a).
\end{equation}

The optimal action-value function obeys the {\it Bellman equation},
\begin{equation}
Q^*(s,a) = \mathbb{E} [r+\gamma \mathop{max}_{a'}Q^*(s',a')|s,a],
\end{equation}
where $s'$ is any sequence derived by taking $a$ after seeing $s$. 

\subsection{Deep Q-Network with Experience Replay}

Deep Q-Network (DQN) uses a neural network to approximate the optimal action-value function $Q^*(s,a)$.  The network is trained by minimizing a sequence of loss functions at each iteration $i$, as
\begin{equation}
L_i(\theta_i) = \mathop{\mathbb{E}}_{s,a \thicksim \rho(\cdot)} \left[\left(y_i - Q(s,a;\theta_i)\right)^2\right]
\end{equation}
where $y_i = \mathbb{E}_{s' \thicksim \mathcal{E}} [r+\gamma \mathop{max}_{a'}Q(s',a';\theta_{i-1})|s,a]$ is a boosted target for the iteration $i$, and $\rho(s,a)$ is a probability distribution over $s$ and $a$ referred as behavior distribution. Differentiating the loss functions with respect to the weights leads to,
\begin{equation}
\nabla_{\theta_i} L_i( \theta_i) = \mathop{\mathbb{E}}_{\footnotesize
 \begin{array}{c}
  s, a \thicksim \rho(.) \\
	s' \thicksim \mathcal{E} \end{array}
} \left[ - \left(r+ \gamma \mathop{max}_{a'}(s',a';\theta_{i-1}) - Q(s,a;\theta_i) \right) \nabla_{\theta_i}Q(s,a;\theta_i) \right].
\end{equation}

DQN simplifies the computation through replacing the expection by single samples from the $\rho$ and $\mathcal{E}$, forumlated as,
\begin{equation}
\nabla_{\theta_i} L_t( \theta_i) = -\left(r_t+\gamma \mathop{max}_{a'}Q(s_{t+1},a';\theta_{i-1}) - Q(s_t,a_t;\theta_i)\right) \nabla_{\theta_i}Q(s_t,a_t;\theta_i). \label{eq:dqn}
\end{equation}

\cite{mnih2013play} proposed utilizing an experience replay technique in DQN, presented by the algorithm~\ref{alg:replay}. The approach stores the agent's experiences at each time-step  formulated as $e_t=(s_t,a_t,r_t,s_{t+1})$,   into a replay memory as $\mathcal{D}=e_1,\ldots,e_N$. The approach then picks random samples from  $\mathcal{D}$ for updating the neural network. 

DQN with experience replay is dramatically more efficient and stable than the standard online Q-learning~\cite{sutton1998reinforcement}. The reasons are as follows~\cite{mnih2013play}.
\begin{itemize}
\item Learning directly from consecutive samples is inefficient due to the strong correlations between the samples; randomizing the samples breaks these correlations and therefore reduces the variance of updates.
\item When learning on-policy, the current parameters determine the next data sample that the parameters are trained on; By using experience replay the behavior distribution is averaged over many of its previous states, smoothing out learning and voiding oscillations or divergence in the parameters.
\item Each step of experience is potentially used in many weight updates, which allows for greater data efficiency.
\end{itemize}

\begin{algorithm} 
\caption{Deep Q-Network with Experience Replay}   
\label{alg:replay}  
{
\begin{algorithmic}[1]   

\State Initialize the replay memory $\mathcal{D}$ to capacity $N$

\State Initialize the action-value function $\mathcal{Q}$ with random weights

\For{episode = $1$, $M$}  

	\State Initialize sequence $s_1 = \{ x_1 \} $ and preprocessed $\phi_1 = \phi (s_1)$

	\For{$t$ = $1$, $T$}

		\State With probability $\epsilon$ select a random action $a_t$
		\State otherwise select $a_t = \mathop{argmax}_a Q^*(\phi(s_t),a;\theta)$ 
		\State Execute action $a_t$ in emulator and observe reward $r_t$ and image $x_{t+1}$
		\State Set $s_{t+1}=s_t,a_t,x_{t+1}$ and preprocess $\phi_{t+1}=\phi(s_{t+1})$
		\State Store transition $(\phi_t, a_t, r_t,\phi_{t+1})$ in $\mathcal{D}$
		\State Sample random minibatch of transitions $(\phi_j, a_j, r_j, \phi_{j+1})$ from $\mathcal{D}$
		\State Set $y_j = \left\{ \begin{array}{ll}
				r_j & \textrm{for terminal} \phi_{j+1} \\
				r_j+\gamma \mathop{max}_{a'}Q(\phi_{j+1}, a'; \theta)  & \textrm{for non-terminal} \phi_{j+1} \\
				\end{array} \right.$
		\State Perform a gradient descent step on $(y_j-Q(\phi_j a_j;\theta))^2$ using the equation~\ref{eq:dqn}
	\EndFor

\EndFor
\end{algorithmic}  
}
\end{algorithm}  

\subsection{Double Deep Q-Network}

\cite{hasselt2015deep} proposed double DQN to reduce the over-estimations caused by the max operation in the equation~\ref{eq:dqn}.

The standard DQN used a training target as,
\begin{equation}
Y_t^{DQN} = r_t+\gamma \mathop{max}_a Q(s_{t+1}, a ; \theta_t^-), \label{eq:dqn:target}
\end{equation}
where $\theta_t^-$ is the parameters of a target network which is copied periodically from the online network. Because DQN is a kind of boosting algorithm, the estimated target $Q(s_{t+1},a; \theta_t^-)$ is unavoidably inaccurate as an oracle function during the training procedure. This inaccuracy is high likely to be converted into over-estimations by the max operation. 

Double DQN decomposes the max operation in the equation~\ref{eq:dqn:target} into action selection and action evaluation, formulated as,
\begin{equation}
Y_t^{DoubleDQN} = r_t+\gamma Q(s_{t+1}, argmaxQ(s_{t+1},a;\theta_t), \theta_t^-),
\end{equation}
where the online network with the parameters $\theta_t$ is used to evaluate the greedy policy, and the target network with the parameters $\theta_t^-$ is used to estimate its values.

\subsection{Prioritized Experience Replay}

\cite{schaul2015prioritized} proposed prioritized experience replay to improve the learning efficiency of DQN, presented by the algorithm~\ref{alg:prioritized}. The intuition of the method is to replay important transitions more frequently.

The probability of sampling a transition $i$ is defined as,
\begin{equation}
P(i) = \frac{p_i ^ \alpha} {\sum_k p_k ^ \alpha}
\end{equation}
where $p_i > 0$ is the priority  of the transition $i$. The exponent $\alpha$ determines how much prioritization is used, with $\alpha=0$ corresponding to the uniform case. 

The priority $p_i$ in the proportional prioritization method,  which is implemented in CytonRL, is defined as,
\begin{equation}
p_i = | \delta_i | + \epsilon, 
\end{equation}
where $\delta_i$ is the prediction error, and $\epsilon$ is a small positive constant that prevents the edge-case of transitions not being revisited once their error is zero.

Prioritized experience replay changes the sampling distribution, which brings bias to the estimation. Importance-sampling is used to compensate this bias, formulated as
\begin{equation}
w_i=(\frac{1}{N} \cdot \frac{1}{P(i)})^\beta
\end{equation}
where $0 \leqslant \beta \leqslant 1 $ controls the strength of compensation.

\begin{algorithm} 
\caption{Double DQN with Proportional Prioritized Experience Replay}   
\label{alg:prioritized}  
{
\begin{algorithmic}[1]   
\Require minibatch $k$, step-size $\eta$, replay period $K$ and size $N$, exponents $\alpha$ and $\beta$, budget $T$.

\State Initialize replay memory $\mathcal{H}$ to capacity $N$, $\Delta=0$, $p_1=1$.

\State Observe $s_1$

\For{t = $1$, $T$}  
	\State Choose action $a_t \thicksim \pi_{\theta}(s_t)$

	\State Observe $r_t$ and $s_{t+1}$
	
	\State Store transition $(s_t, a_t, r_t, s_{t+1})$ in $\mathcal{H}$ with maximal priority $p_i = max_{i<t} \; p_i$
	\If{$t = 0$  mod $K$}

		\For{$j$ = $1$, $k$}

			\State Sample transition $j \thicksim P(j) = p_j^\alpha / \sum_i p_i^\alpha$
			\State Compute importance-sampling weight $w_j = (N \cdot P(j))^{-\beta}/max_i \; w_i$
			\State Compute TD-error $\delta_j = r_j + \gamma Q_\mathrm{target} (s_{j+1}, \mathop{argmax}_a Q(s_j,a)) - Q(s_j, a_j)$
			\State Update transition priority $p_j \gets |\delta_j|$
			\State Accumulate weight-change $\Delta \gets \Delta + w_j \cdot \delta_j \cdot \nabla_\theta Q(s_j, a_j)$
		\EndFor
		\State Update weights $\theta \gets \theta + \eta \cdot \Delta$
		\State $\Delta \gets 0$.
		\State From the time to time copy wiehgts into target network $\theta_\textrm{target} \gets \theta$
	\EndIf

\EndFor
\end{algorithmic}  
}
\end{algorithm}  

\subsection{Dueling DQN}

\cite{wang2015dueling} proposed an dueling neural network architecture for DQN, which decomposed the Q function into two separate estimators; one for the state-value function and one for the state-dependent action-advantage function.

The Q function of dueling DQN is formulated as,
\begin{eqnarray}
Q(s,a; \theta, \alpha, \beta) & = & V(s; \theta, \beta) +
\left (A(s,a;\theta,\alpha) - \frac{1}{|A|}\sum_{a'}A(s,a';\theta,\alpha) \right)
\end{eqnarray}
where $V(s; \theta, \beta)$ is an state-value function, and $A(s,a;\theta,\alpha)$ is an state-dependent action-advantage function. Note that the formula makes that the action-advantage function has no overall impact on the state-value function. In this way, the two functions can be uniquely derived from any Q function.





\section{Implementation}
\label{sec:implementation}

CytonRL is implemented using the C++ language with a dependency on OpenCV~\footnote{https://github.com/opencv/opencv} to down-sample the input images, and a dependency on NVIDIA's GPU-accelerated libraries -- {\it cuda}, {\it cublas} and {\it cudnn} to use GPUs. CytonLib -- a general purpose C++ neural network library -- is shipped together with the toolkit, which greatly reduced the workload of writing C++ codes for GPUs.

\begin{figure}[tb!]
\begin{center}
\vspace{25pt}
\includegraphics[width=1.0\textwidth]{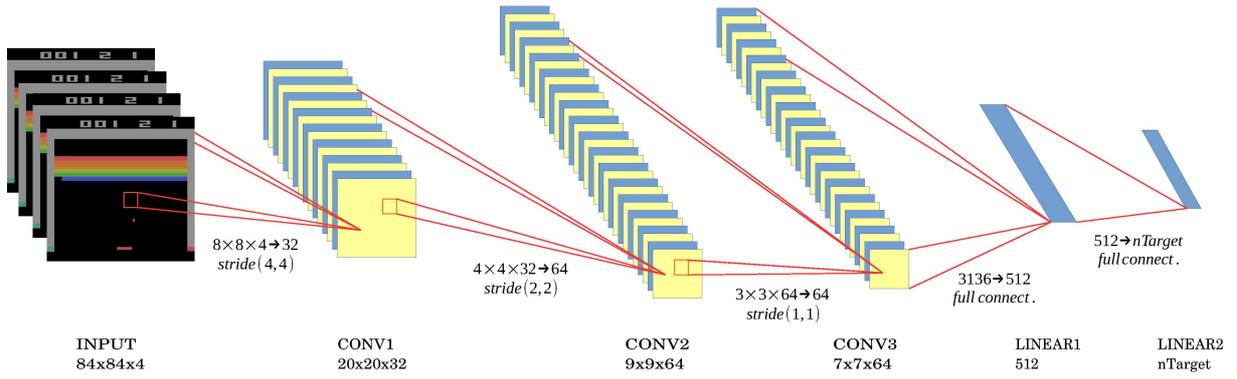}
\end{center}
\caption{Architecure of Neural Network}
\label{fig:arch}
\end{figure}

The neural network architecture used by CytonRL is illustrated by the figure~\ref{fig:arch}, which have been established for the Atari games~\cite{mnih2013play,mnih2015human}. The input of the neural network is processed Atari frames. The raw Atari frames are $210 \times 100$ pixel images with a 128 color palette. The images are converted to grey-scale, and linearly interpreted into $84 \times 84 $ through the OpenCV library. The consecutive 4 images are concatenated to form an $84 \times 84 \times 4$ tensor, which is taken as the input to the neural network. The structures of each layer in the neural network are as follows.
\begin{itemize}
\item The first layer uses convolution connections with a $8 \times 8 \times 4 \mapsto 32 $ filter with a stride of (4, 4).
\item The second layer uses convolution connections with a $4 \times 4 \times 32 \mapsto 64 $ filter with a stride of (2, 2).
\item The third layer uses convolution connections with  a $3 \times 3 \times 64 \mapsto 64 $ filter with a stride of (1,1). 
\item The fourth layer uses full connections with 512 units.
\item The fifth layer uses full connections with the same number of units as the target signals.
\end{itemize}
The activation functions of all layers are rectified linear function~\cite{nair2010rectified}.

The source code that implements the above neural network architecture is presented in the figure~\ref{fig:code:architecture}. The code uses CytonLib to build a fully operable neural network. Note that the code is slightly simplified to emphasize the working mechanism.  The code works as follows,

\begin{itemize}

\item The class of {\it Variable} stores numeric values and gradients. Through passing the pointer of {\it Variable} around, all components are connected.

\item The data member {\it layers}  collects  all the components. The base class of {\it Network} calls the functions {\it forward}, {\it backward} and {\it calculateGradient} of each component to perform the actual computation, illustrated by the figure~\ref{fig:code:network}.

\end{itemize}

\begin{figure}
\begin{center}
\noindent\cprotect\fbox{\begin{minipage}{0.8\textwidth}
{\footnotesize
\renewcommand{\baselinestretch}{0.95}
\begin{verbatim} 
class NetworkRL: public Network
{
  ConvolutionLayer conv1; // declare components
  ActivationLayer act1;
  ConvolutionLayer conv2;
  ActivationLayer act2;
  ConvolutionLayer conv3;
  ActivationLayer act3;
  LinearLayer lin1;
  ActivationLayer act4;
  DueLinearLayer dueLin2;
  LinearLayer lin2;
  
  void init(Variable* x, int nTarget) 
  // x: input of the neural network which is image data
  // nTarget: dimension of output which equals to the number of 
  //          control signals
  {

    tx=conv1.init(tx, 32, 8, 8, 4, 4, 0, 0);
    layers.push_back(&conv1);

    tx=act1.init(tx, CUDNN_ACTIVATION_RELU);
    layers.push_back(&act1);
  
    tx=conv2.init(tx, 64, 4, 4, 2, 2, 0, 0);
    layers.push_back(&conv2);
  
    tx=act2.init(tx, CUDNN_ACTIVATION_RELU);
    layers.push_back(&act2);
  
    tx=conv3.init(tx, 64, 3, 3, 1, 1, 0, 0);
    layers.push_back(&conv3);
  
    tx=act3.init(tx, CUDNN_ACTIVATION_RELU);
    layers.push_back(&act3);
  
    tx=lin1.init(tx, 512, true  );
    layers.push_back(&lin1);
  
    tx=act4.act4(tx, CUDNN_ACTIVATION_RELU);
    layers.push_back(&act4);
    
    if(params.dueling)
    {
      tx=dueLin2.init(tx, nTarget, true);
      layers.push_back(&dueLin2);
    }
    else
    {
      tx=lin2.init(tx, nTarget, true);
      layers.push_back(&lin2);
    }
    
    return tx;  //pointer to result
  }
};
\end{verbatim}
}
\end{minipage}}
\end{center}
\caption{Source Code of Neural Network}
\label{fig:code:architecture}
\end{figure}

\begin{figure}
\begin{center}
\noindent\cprotect\fbox{\begin{minipage}{0.8\textwidth}
{\footnotesize
\renewcommand{\baselinestretch}{0.95}
\begin{verbatim} 
class Layer
{
  virtual void forward(){};

  virtual void backward(){};

  virtual void calculateGradient(){};
};

class Network: public Layer
{

  vector<Layer*> layers;

  void forward()
  {
    for(int k=0; k<layers.size(); k++)
      layers.at(k)->forward();
  }

  void backward()
  {
    for(int k=layers.size()-1; k>=0; k--)
      layers.at(k)->backward();
  }

  void calculateGradient()
  {
    for(int k=layers.size()-1; k>=0; k--)
      layers.at(k)->calculateGradient();
  }
};
\end{verbatim}
}
\end{minipage}}
\end{center}
\caption{Source Code of Network Class}
\label{fig:code:network}
\end{figure}

The code of actual computation is organized in the functions {\it forward}, {\it backward} and {\it calculateGradient} for each type of component. The figure~\ref{fig:code:working} presents some examples. 

\begin{figure}
\begin{center}
\noindent\cprotect\fbox{\begin{minipage}{0.8\textwidth}
{\footnotesize
\renewcommand{\baselinestretch}{0.95}
\begin{verbatim}

void LinearLayer::forward()
{
 cublasXgemm(cublasH, CUBLAS_OP_T, CUBLAS_OP_N,
   dimOutput, num, dimInput,
   &one, w.data, w.ni, x.data, dimInput,
   &zero, y.data, dimOutput)
}

void LinearLayer::backward()
{
 cublasXgemm(cublasH, CUBLAS_OP_N, CUBLAS_OP_N,
   dimInput, num, dimOutput,
   &one, w.data, w.ni, y.grad.data, dimOutput,
   &beta, x.grad.data, dimInput));
}

void LinearLayer::calculateGradient()
{
 cublasXgemm(cublasH, CUBLAS_OP_N, CUBLAS_OP_T,
   dimInput, dimOutput,  num,
   &one, x.data, dimInput, y.grad.data, dimOutput,
   &one, w.grad.data, w.grad.ni));
}


void ActivationLayer::forward()
{
  cudnnActivationForward(global.cudnnHandle, activeDesc,
    &global.one, x->desc, x->data,
    &global.zero, y.desc, y.data) );
}

\end{verbatim}
}
\end{minipage}}
\end{center}
\caption{Source Code of Performing Actual Computation }
\label{fig:code:working}
\end{figure}

\section{Benchmark}
\label{sec:benchmark}

\subsection{Settings}

The hyperparameter settings of CytonRL used in the benchmarks are presented by the table~\ref{tab:hparam}, which are coded as the default settings. The settings are initially based on the \cite{mnih2013play,hasselt2015deep,schaul2015prioritized,wang2015dueling}, and modified to improve the stability and efficiency according to our experiments.

\begin{table}
\begin{center}
\begin{tabular}{|l|l|}
\hline
\bf Hyperparameter & \bf Value \\
\hline
Replay Memory Size & 1,000,000 \\
Input Frames & 4\\
$\gamma$               & 0.99 \\
Learning Rate & 0.000625\\
Prioritized Exp. Replay $\alpha$ & 0.6  \\
Prioritized Exp. Replay $\beta$ &  0.4 $\to$ 1.0 (1 $\to$ Maximum Training Step) \\
Train. $\epsilon$-greedy  & 1.0 $\to$ 0.1 (1 $\to$ 5,000,000 steps) \\
Test $\epsilon$-greedy & 0.001 \\
Learning Start & 50,000 steps \\
Batch Size & 32 \\
Update Period &  4 steps \\
TargetQ update & 30,000 steps \\
Maximum Training Step & 100,000,000 steps \\
Maximum Step per Episode & 18,000 steps\\
Test Period & 5,000,000 steps\\
Optimizer & RMSprop \\
\hline
\end{tabular}
\end{center}
\caption{\label{tab:hparam}Hyperparameter Settings}
\end{table}

\subsection{Performance}

\begin{figure}[tb!]
\begin{center}
\vspace{25pt}
\includegraphics[width=0.8\textwidth]{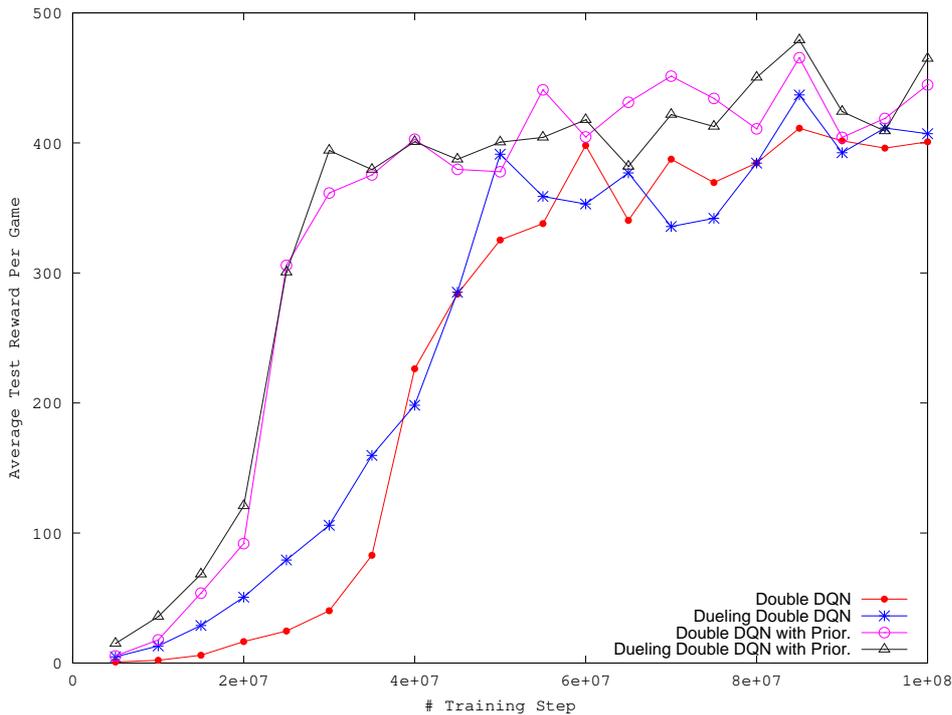}
\end{center}
\caption{Performance on the Atari Game of Breakout}
\label{fig:performance}
\end{figure}

The performance of CytonRL on the popular Atari game of Breakout is presented in the figure~\ref{fig:performance}.  CytonRL was run with four model settings as,
\begin{description}
\item[double DQN] : --dueling 0 --priorityAlpha 0
\item[dueling double DQN] : --dueling 1 --priorityAlpha 0
\item[double DQN with prioritized replay] : --dueling 0 --priorityAlpha 0.6
\item[dueling double DQN with prioritized replay] : --dueling 1 --priorityAlpha 0.6
\end{description}
For echo model setting, CytonRL was trained 100,000,000 steps, and tested very 5,000,000 steps. In each test, 100 games were played, and the rewards of all games were averaged. The curves of training steps versus average test reward per game are presented in the figure.

The results reveal the strength of each model settings as {\it dueling double DQN with prior.} $\succ$ {\it double DQN with prior.} $\succ$ {\it dueling double DQN} $\succ$ {\it double DQN}. Stronger model settings tend to learn the game faster, and achieve better final performance.  The results confirm that {\it DQN}, {\it double DQN}, {\it prioritized relay}, and {\it dueling DQN} are all effective RL algorithms.

\section{Conclusion}
\label{sec:conclusion}

This paper introduces CytonRL -- an open-source reinforcement learning toolkit built from scratch using C++ and NVIDA's GPU-accelerated libraries. CytonRL is coded and tuned to achieve competitive performances in a fast manner. In other words, the toolkit is both effective and efficient.  The source code of CytonRL is simple because of CytonLib -- an open-source general purpose neural network library – which is contained in the toolkit.  Therefore, CytonRL is an attractive alternative choice for the research community. We open-source this toolkit in the hope to benefit the community and promote the field. We look forward to hearing feedback.

\bibliography{ref}
\bibliographystyle{acl_natbib}

\end{document}